\documentclass[runningheads]{llncs}
\usepackage{graphicx}

\usepackage{comment}
\usepackage{amsmath,amssymb} 
\usepackage{color}
\usepackage{dsfont}
\usepackage{booktabs}
\usepackage{xcolor}
\usepackage{colortbl}
\usepackage{multirow}
\usepackage{paralist}
\usepackage{marvosym}
\usepackage{algorithm,algorithmicx,algpseudocode}
\usepackage{url}

\usepackage{colortbl}
\usepackage{hhline}
\definecolor{mygray}{gray}{.9}

\begin{document}
\pagestyle{headings}
\mainmatter

\def\ACCV20SubNumber{591}  

\title{Synthesizing the Unseen for Zero-shot Object Detection} 
\titlerunning{Synthesizing the Unseen for ZSD}
%
\author{Nasir Hayat\inst{1} \and
Munawar Hayat\inst{1,2} \and
Shafin Rahman\inst{3} \and
Salman Khan\inst{1,2} \\
Syed Waqas Zamir \inst{1} \and
Fahad Shahbaz Khan \inst{1,2} 
}
\authorrunning{N. Hayat et al.}
%
\institute{Inception Institute of Artificial Intelligence, UAE\and
MBZ University of AI, UAE \quad \quad $^3\hspace{1mm} $North South University, Bangladesh
\email{nh2218@nyu.edu,\{munawar.hayat, salman.khan, fahad.khan\}@mbzuai.ac.ae}}



\maketitle


\begin{abstract}
The existing zero-shot detection approaches project visual features to the semantic domain for seen objects, hoping to map unseen objects to their corresponding semantics during inference. However, since the unseen objects are never visualized during training, the detection model is skewed towards seen content, thereby labeling unseen as background or a seen class. In this work, we propose to \emph{synthesize} visual features for unseen classes, so that the model learns both seen and unseen objects in the visual domain. Consequently, the major challenge becomes, \emph{how to accurately synthesize unseen objects merely using their class semantics?} Towards this ambitious goal, we propose a novel generative model that uses class-semantics to not only generate the features but also to discriminatively separate them. Further, using a unified model, we ensure the synthesized features have high diversity that represents the intra-class differences and variable localization precision in the detected bounding boxes. We test our approach on three object detection benchmarks, PASCAL VOC, MSCOCO, and ILSVRC detection, under both conventional and generalized settings, showing impressive gains over the state-of-the-art methods. Our codes are available at \url{https://github.com/nasir6/zero_shot_detection}
\end{abstract}

\section{Introduction}

Object detection is a challenging  problem that seeks to simultaneously localize and classify object instances in an image \cite{ren2015faster}. Traditional object detection methods work in a supervised setting where a large amount of annotated data is used to train models. Annotating object bounding boxes for training such models is a labor-intensive and expensive process. Further, for many rare occurring objects, we might not have any training examples. Humans, on the other hand, can easily identify unseen objects solely based upon the objects' attributes or their natural language description. Zero Shot Detection (ZSD) is a recently introduced paradigm which enables simultaneous localization and classification of previously \emph{unseen} objects. It is arguably the most extreme case of learning with minimal supervision  
\cite{rahman2018zero,bansal2018zero}.

ZSD is commonly accomplished by learning to project visual representations of different objects to a pre-defined semantic embedding space,  
and then performing nearest neighbor search in the semantic space at inference \cite{rahman2018zero,bansal2018zero,demirel2018zero,li2019zero}. 
Since the unseen examples are never visualized during training, the model gets significantly biased towards the seen objects \cite{hayat2019gaussian,khan2019striking}, leading to problems such as confusion with background and mode collapse resulting in high scores for only some unseen classes. In this work, we are motivated by the idea that if an object detector can visualize the unseen data distribution, the above-mentioned problems can be alleviated. To this end, we propose a conditional feature generation module to synthesize visual features for unseen objects, that are in turn used to directly adapt the classifier head of Faster-RCNN \cite{ren2015faster}. While such feature synthesis approaches have been previously explored in the context of zero-shot classification, they cannot be directly applied to ZSD due to the unique challenges in detection setting such as localizing multiple objects per image and modeling diverse backgrounds. 




The core of our approach is a novel feature synthesis module, guided by semantic space representations, which is capable of generating diverse and discriminative visual features for unseen classes. 
We generate exemplars in the feature space and use them to modify the projection vectors corresponding to unseen classes in the Faster-RCNN classification head. The major contributions of the paper are: \textbf{(i)} it proposes a novel approach to visual feature synthesis conditioned upon class-semantics and regularized to enhance feature diversity, \textbf{(ii)} feature generation process is jointly driven by classification loss in the semantic space for both seen and unseen classes, to ensure that generated features are discriminant and compatible with the object-classifier, \textbf{(iii)} extensive experiments on Pascal VOC, MSCOCO and ILSVRC detection datasets to demonstrate the effectiveness of the proposed method. For instance, we achieve a relative mAP gain of $53\%$ on MS-COCO dataset over existing state-of-the-art on ZSD task. Our approach is also demonstrated to work favorably well for Generalized ZSD (GZSD) task that aims to detect both \emph{seen} and \emph{unseen} objects.

\section{Related Work}
\noindent\textbf{Zero-shot Recognition:}
The goal of Zero shot learning (ZSL) is to classify images of unseen classes given their textual semantics in the form of wordvecs \cite{zhang2017learning}, text-descriptions \cite{lei2015predicting,li2019zero} or human annotated attributes \cite{annadani2018preserving}. This is commonly done by learning a joint embedding space where semantics and visual features can interact. The embeddings can be learnt to project from visual-to-semantic \cite{lampert2013attribute}, or semantic-to-visual space \cite{zhang2017learning}. Some methods also project both visual and semantic features into a common space \cite{akata2015evaluation}. The existing methods which learn a projection or embedding space have multiple inherent limitations such as the hubness problem \cite{dinu2014improving} caused by shrinked low dimensional semantic space with limited or no diversity to encompass variations in the visual image space. These methods are therefore prone to mis-classify unseen samples into seen due to non-existence of training samples for the unseen. 
Recently, generative approaches deploying variational auto-encoders (VAEs) or generative adverserial networks (GANs) have shown promises for ZSL \cite{chen2018zero,xian2018feature,zhu2018generative,khan2018adversarial}. These approaches model the underlying data distribution of visual feature space by training a generator and a discriminator network that compete in a minimax game, thereby synthesizing features for unseen classes conditioned on their semantic representations.

\noindent\textbf{Zero-shot Detection:}
The existing literature on zero shot learning is dominated by zero shot classification (ZSC). Zero Shot Detection (ZSD), first introduced in \cite{rahman2018zero,bansal2018zero}, is significantly more challenging compared with ZSC, since it aims to simultaneously localize and classify an unseen object. \cite{rahman2018zero} maps visual features to a semantic space and enforces max-margin constraints along-with meta-class clustering to enhance inter-class discrimination. The authors in \cite{bansal2018zero} incorporate an improved semantic mapping for the background in an iterative manner by first projecting the seen class visual features to their corresponding semantics and then the background bounding boxes to a set of diverse unseen semantic vectors. \cite{demirel2018zero} learns an embedding space as a convex combination of training class wordvecs. \cite{li2019zero} uses a Recurrent Neural Network to model natural language description of objects in the image.


Unlike ZSC, synthetic feature generation for unseen classes is less investigated for ZSD and only \cite{zhao2020gtnet} augments features. Ours is a novel feature synthesis approach that has the following major differences from \cite{zhao2020gtnet} \textbf{(i)} For feature generation, we only train a single GAN model, in comparison to \cite{zhao2020gtnet} which trains three isolated models. Our unified GAN model is capable of generating diverse and distinct features for unseen classes. \textbf{(ii)} We propose to incorporate a semantics guided loss function, which improves feature generation capability of the generator module for unseen categories. \textbf{(iii)} To enhance diversification amongst the generated features, we incorporate a mode seeking regularization term. We further compare our method directly with \cite{zhao2020gtnet} and show that it outperforms \cite{zhao2020gtnet} by a significant margin, while using a single unified generation module.



\section{Method}

\textbf{Motivation:} Most of the existing approaches for ZSD address this problem in the semantic embedding space. This means that the visual features are mapped to semantic domain where unseen semantics are related with potential unseen object features to predict decision scores.  We identify three problems with this line of investigation. \textbf{(i)} \emph{Unseen background confusion:} Due to the low objectness scores for unseen objects, they frequently get confused as background during inference. To counter this, \cite{bansal2018zero,rahman2018polarity} use external data in the form of object annotations or vocabulary that are neither seen nor unseen. \textbf{(ii)} \emph{Biasness problem:} Since, unseen objects are never experienced during training, the model becomes heavily biased towards seen classes. For this, approaches usually design specialized loss functions to regularize learning \cite{rahman2018polarity,rahman2018zero}. \textbf{(iii)} \emph{Hubness problem:} Only a few unseen classes get the highest scores in most cases. Addressing the problem in semantic space intensifies the hubness issue \cite{Zhang_2017_CVPR}. Very recently, GTNet \cite{zhao2020gtnet} attempted to address these issues in the visual domain instead of the semantic space. Similar to \cite{xian2018feature}, they generate synthesized features to train unseen classes in a supervised manner. We identify two important drawbacks in this approach. \textbf{(i)} They train multiple GAN models to incorporate variance due to intra-class differences and varying overlaps with ground-truth (IoU). These generative models are trained in a sequential manner, without an end-to-end learning mechanism, making it difficult to fix errors in early stages.  
\textbf{(ii)} In addition to synthesized unseen object features, they need to generate synthesized background features. As the background semantic is not easy to define, synthesized background features become too noisy than that of object features, thereby significantly hindering the learning process. In this paper, we attempt to solve this problem by training one unified GAN model to generate synthesized unseen object features that can be used to train with real background features without the help of synthesized background features. Further, without requiring multiple sequential generative models to inject feature diversity \cite{zhao2020gtnet}, we propose a simple regularization term to promote diversity in the synthesized features.

\subsection{Overview}
\noindent \textbf{Problem Formulation:} Consider the train set $ \mathcal{X}^{s}$  contains image of seen objects and the test set  $\mathcal{X}^{u}$ contains images of seen+unseen objects. Each image can have multiple objects.  Let's denote $\mathcal{Y}_s =\{1,\cdots S\}$ and $\mathcal{Y}_u =\{S+1, \cdots S+U\}$ respectively as the label sets for seen and unseen classes. Note that $S$ and $U$ denote total number of seen and unseen classes respectively, and $\mathcal{Y}_S \cap \mathcal{Y}_u = \emptyset$. At training, we are given annotations in terms of class labels $y \in \mathcal{Y}_s$ and bounding-box coordinates $b \in \mathbb{R}^{4}$ for all seen objects in $ \mathcal{X}^{s}$. We are also given semantic embeddings $\mathbf{W}_s \in \mathbb{R}^{d\times S}$ and $\mathbf{W}_u \in \mathbb{R}^{d\times U}$ for seen and unseen classes respectively (e.g., Glove \cite{pennington2014glove} and fastText \cite{joulin2017bag}). 
At inference, we are required to correctly predict the class-labels and bounding-box coordinates for the objects in images of $\mathcal{X}^{u}$. For ZSD settings, only unseen predictions are required, while for generalized ZSD, both seen and unseen predictions must be made.

\begin{figure}[tp]
    \centering
    \includegraphics[width=\textwidth, clip=true, trim=0cm 7.3cm 0cm 0cm]{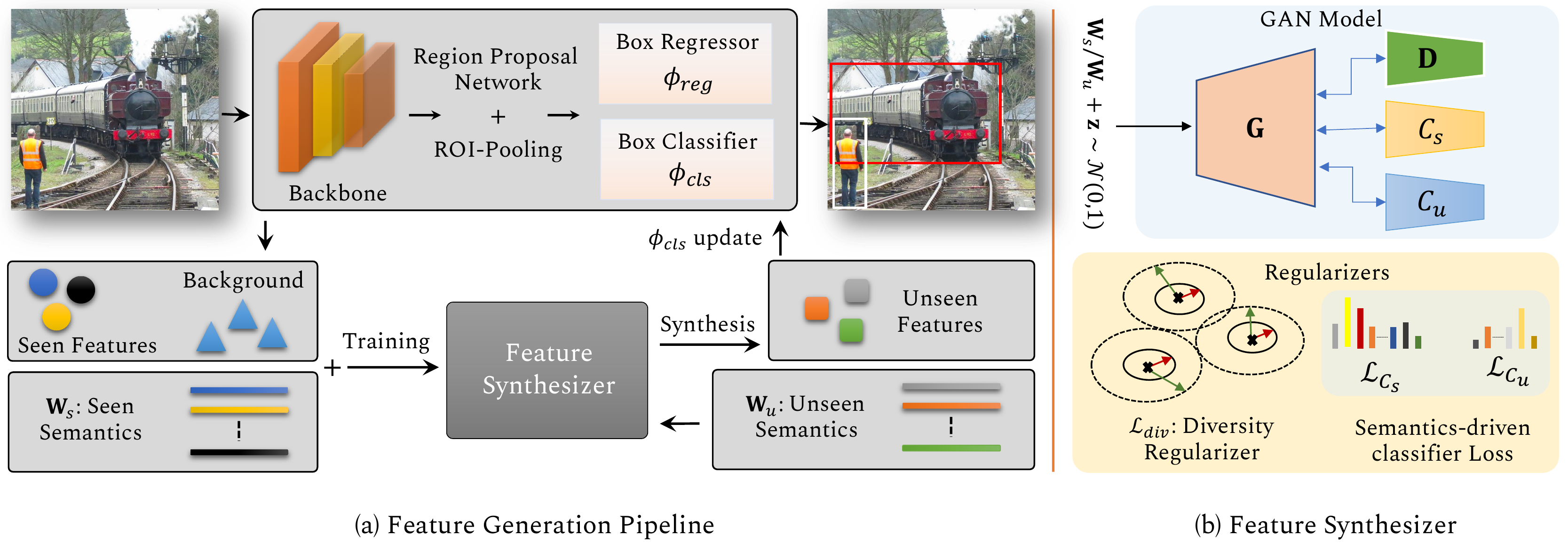}
    \caption{Overview of proposed generative ZSD approach.}
    \label{fig:main_fig}
\end{figure}

We outline different steps used for our generative ZSD pipeline in Alg.~\ref{alg: main} and Fig.~\ref{fig:main_fig} illustrates our method. The proposed ZSD framework is designed to work with any two-stage object detector. For this paper, we implement Faster-RCNN model with ResNet-101 backbone. We first train the Faster-RCNN model $\phi_{\texttt{faster-rcnn}}$ on the training images $ \mathcal{X}^{s}$ comprising of only seen objects and their corresponding ground-truth annotations. Given an input image $\mathbf{x} \in \mathcal{X}^{s}$, it is first represented in terms of activations of a pre-trained ResNet-101. Note that the backbone ResNet-101 was trained on ImageNet data by \emph{excluding} images belonging to the overlapping unseen classes of the evaluated ZSD datasets. The extracted features are feed-forwarded to the region proposal network (RPN) of Faster-RCNN, which generates a set of candidate object bounding box proposals at different sizes and aspect ratios. These feature maps and the proposals are then mapped through an RoI pooling layer, to achieve a fixed-size representation for each proposal. Let's denote the feature maps corresponding to $K$ bounding box proposals of an image with $\mathbf{f}_i \in \mathbb{R}^{1024}, i=1,\cdots K$. The features $\mathbf{f}_i$ are then passed through two modules: bounding-box-regressor, and object-classifier. Once $\phi_{\texttt{faster-rcnn}}$ is trained on the seen data $ \mathcal{X}^{s}$, we use it to extract features for seen object anchor boxes. All candidate proposals with an intersection-over-union (IoU) $\ge 0.7$ are considered as foreground, whereas the ones with IoU $\le 0.3$ are considered backgrounds. For $N_{tr}$ training images in $\mathcal{X}^{s}$, we therefore get bounding-box features $\mathbf{F}_s \in \mathbb{R}^{1024 \times K.N_{tr}}$ and their class-labels $\mathbf{Y}_s \in \mathbb{R}^{K.N_{tr}}$. Next, we learn a unified generative model to learn the relationship between visual and semantic domains.

\begin{algorithm*}[!t]
  \caption{The proposed  feature synthesis base ZSD method}
  \label{alg}
  \begin{algorithmic}[1]
  \Require $\mathcal{X}^s,\mathcal{X}^u, y \in \mathcal{Y}_s, b, \mathbf{W}_s, \mathbf{W}_u$
\State $\phi_{\texttt{faster-rcnn}} \leftarrow$ Train Faster-RCNN using seen data $\mathcal{X}^s$ and annotations
\State $\mathbf{F}_s,\mathbf{Y}_s \leftarrow$ Extract features for b-boxes of $\mathcal{X}^s$ using RPN of $\phi_{\texttt{faster-rcnn}}$
\State $\mathbf{\phi}_{\texttt{Ws-cls}} \leftarrow$ Train $\mathbf{\phi}_{\texttt{Ws-cls}}$ using $\mathbf{F}_s,\mathbf{Y}_s$
\State $\mathbf{\phi}_{\texttt{Wu-cls}} \leftarrow$ Define $\mathbf{\phi}_{\texttt{Wu-cls}}$ using $\mathbf{\phi}_{\texttt{Ws-cls}}$ by replacing $\mathbf{W}_s$ with $\mathbf{W}_u$
\State $\mathbf{G}\leftarrow$ Train GAN by optimizing loss in Eq.~\ref{eqn: full_gan_loss}
\State $\mathbf{\tilde{F}}_u, \mathbf{Y}_u \leftarrow$ Syntesize features for unseen classes using $\mathbf{G}$ and $\mathbf{W}_u$
\State $\mathbf{\phi}_{\texttt{cls}}^\prime \leftarrow$ Train $\mathbf{\phi}_{\texttt{cls}}$ using $\mathbf{\tilde{F}}_u, \mathbf{Y}_u$
\State $\phi_{\texttt{faster-rcnn}\leftarrow}$ Update $\phi_{\texttt{faster-rcnn}}$ with $\mathbf{\phi}_{\texttt{cls}}^\prime$
\State Evaluate $\phi_{\texttt{faster-rcnn}}$ on $\mathcal{X}^u$
\Ensure Class labels and bbox-coordinates for $\mathcal{X}^u$
  \end{algorithmic}
  \label{alg: main}
\end{algorithm*}

\subsection{Unified Generative Model}

Given object features $\mathbf{F}_s$, their class-labels $\mathbf{Y}_s$, and semantic vectors $\mathbf{W}_s$ for seen training data $ \mathcal{X}^{s}$, our goal is to learn a conditional generator $\mathbf{G}: \mathcal{W} \times \mathcal{Z} \mapsto\! \mathcal{F}$, which takes a class embedding $\mathbf{w} \in \mathcal{W}$ and a random noise vector $\mathbf{z} \sim \mathcal{N} (\mathbf{0},\mathbf{1}) \in  \mathbb{R}^d$ sampled from a Gaussian distribution and outputs the features $\mathbf{\tilde{f}} \in \mathcal{F}$. The generator $\mathbf{G}$ learns the underlying distribution of the visual features $\mathbf{F}_s$ and their relationship with the semantics $\mathbf{W}_s$. Once trained, the generator $\mathbf{G}$ is used to generate unseen class visual features. Specifically, our feature generation module optimizes the the following objective function,
\begin{equation}
    \min_{\mathbf{G}} \max_{\mathbf{D}} \alpha_{1} \mathcal{L}_{\texttt{WGAN}} +\alpha_{2} \mathcal{L}_{C_s} +\alpha_{3} \mathcal{L}_{C_u} + \alpha_4 \mathcal{L}_{\texttt{div}},
    \label{eqn: full_gan_loss}
\end{equation}
\noindent where $\mathcal{L}_{\texttt{WGAN}}$ minimizes the Wasserstein distance, conditioned upon class semantics,  $\mathcal{L}_{C_s}$ ensures the seen class features generated by $\mathbf{G}$ are suitable and aligned with a pre-trained classifier $\mathbf{\phi}_{\texttt{cls}}$, and $\mathcal{L}_{C_u}$ ensures the synthesized features for unseen classes are aligned with their semantic representations $\mathbf{W}_u$. $\alpha_{1},\alpha_{2},\alpha_{3},\alpha_{4}$ are the weighting hyper-parameters optimized on a held-out validation set.  The proposed approach is able to generate sufficiently discriminative visual features to train the softmax classifier. Each term in Eq.~\ref{eqn: full_gan_loss} is discussed next.

\subsection{Conditional Wasserstein GAN}
We build upon improved WGAN \cite{gulrajani2017improved} and extend it to conditional WGAN (cWGAN), by integrating the class embedding vectors. The loss $\mathcal{L}_{\texttt{WGAN}}$ is given by,
\begin{equation}
    \mathcal{L}_{\texttt{WGAN}} = \mathbb{E}[\mathbf{D}(\mathbf{f},y)] - \mathbb{E}[\mathbf{D}(\tilde{\mathbf{f}},y)] + \lambda \mathbb{E}[( || \nabla_{\hat{\mathbf{f}}} \mathbf{D} (\hat{\mathbf{f}},y) ||_2 -1 ) ^2],
\end{equation}
\noindent where $\mathbf{f}$ are the real visual features, $\tilde{\mathbf{f}} = \mathbf{G}(\mathbf{w},\mathbf{z})$ denotes the synthesized visual features conditioned upon class semantic vector $\mathbf{w} \in \mathbf{W}_s$, $\hat{\mathbf{f}}= \alpha \mathbf{f} + (1-\alpha) \tilde{\mathbf{f}}$, $\alpha \sim \mathcal{N}(0,1)$ and $\lambda$ is the penalty coefficient. The first two terms provide an approximation of the Wasserstein distance, while the third term enforces gradients to a unit norm along the line connecting pairs of real and generated features.

\subsection{Semantically Guided Feature Generation}
Our end goal is to augment visual features using the proposed generative module such that they enhance discrimination capabilities of the classifier $\mathbf{\phi}_{\texttt{cls}}$. In order to encourage the synthesized features $\tilde{\mathbf{f}} = \mathbf{G}(\mathbf{w},\mathbf{z})$ to be meaningful and discriminative, we optimize the logliklihood of predictions for synthesized seen-class features,
\begin{equation}
    \mathcal{L}_{C_s} = - \mathbb{E}[\log p(y|\mathbf{G}(\mathbf{w},\mathbf{z}); \mathbf{\phi}_{\texttt{cls}})], \quad s.t., \mathbf{w} \in \mathbf{W}_s,
    \label{eqn: lcs}
\end{equation}
where, $y \in \mathcal{Y}_s$ denotes the ground-truth seen class labels, and $p(y|\mathbf{G})$ is the class prediction probability computed by the linear softmax classifier $\mathbf{\phi}_{\texttt{cls}}$. Note that $\mathbf{\phi}_{\texttt{cls}}$ was originally trained on the seen data $\mathcal{X}^s$ and is kept frozen for the purpose of computing $\mathcal{L}_{C_s}$. While the conditional Wasserstein GAN captures underlying data distribution of visual features, the $\mathcal{L}_{C_s}$ term enforces additional constraint and acts as a regularizer to enforce the generated features to be discriminative.

The $\mathcal{L}_{C_s}$ term in Eq.~\ref{eqn: lcs} can act as a regularizer for seen classes only. This is because $\mathcal{L}_{C_s}$ employs pre-trained $\mathbf{\phi}_{\texttt{cls}}$ which was learnt for seen data. In order to enhance the generalization capability of our generator $\mathbf{G}$ towards unseen classes, we propose to incorporate another loss term $\mathcal{L}_{C_u}$.
For this purpose, we redefine the classifier head in terms of class semantics, as $\mathbf{\phi}_{\texttt{Ws-cls}}: \mathbf{f} \xrightarrow{} \texttt{fc} \xrightarrow{} \mathbf{W}_s \xrightarrow{} \texttt{softmax} \xrightarrow{} y_{pr}$, where $\mathbf{f} \in \mathbb{R}^{1024}$ are the input features, $\texttt{fc}$ is the learnable fully-connected layer with weight matrix $\mathbf{W}_{\texttt{fc}} \in \mathbb{R}^{1024 \times d}$ and bias $\mathbf{b}_{\texttt{fc}} \in \mathbb{R}^{d}$, $\mathbf{W}_s \in \mathbb{R}^{d\times S}$ are the fixed non-trainable seen class semantics. The outputs of \texttt{fc} layer are matrix multiplied with $\mathbf{W}_s$ followed by softmax operation to compute class predictions $y_{pr}$. The classifier $\mathbf{\phi}_{\texttt{Ws-cls}}$ is trained on the features $\mathbf{F}_s$ and ground-truth labels $\mathbf{Y_s}$ of seen class bounding boxes. We can then easily define an unseen classifier $\mathbf{\phi}_{\texttt{Wu-cls}}$ by replacing the semantics matrix $\mathbf{W}_s$ in $\mathbf{\phi}_{\texttt{Ws-cls}}$  with $\mathbf{W}_u$.
The semantics guided regularizer loss term $\mathcal{L}_{C_u}$ for synthesized unseen samples is then given by,
\begin{equation}
    \mathcal{L}_{C_u} = - \mathbb{E}[\log p(y|\mathbf{G}(\mathbf{w},\mathbf{z}); \mathbf{\phi}_{\texttt{Wu-cls}})], \quad s.t., \mathbf{w} \in \mathbf{W}_u.
\end{equation}

The $\mathcal{L}_{C_u}$ term therefore incorporates the unseen class-semantics information into feature synthesis, by ensuring that unseen features, after being projected onto \texttt{fc} layer are aligned with their respective semantics vectors.

\subsection{Enhancing Synthesis Diversity} 
Variations in synthesized features are important for learning a robust classifier. Our cWGAN based approach maps a single class semantic vector to multiple visual features. We observed that the conditional generation approach can suffer from mode collapse \cite{salimans2016improved} and generate similar output features conditioned upon prior semantics only, where the noise vectors (responsible for variations in the generated features) get ignored. In order to enhance the diversity of synthesized features, we adapt the mode seeking regularization which maximizes the distance between generations with respect to their corresponding input noise vectors \cite{mao2019mode}. For this purpose, we define the diversity regularization loss $\mathcal{L}_{\texttt{div}}$ as,
\begin{equation}
    \mathcal{L}_{\texttt{div}} = \mathbb{E} [ || \mathbf{G}(\mathbf{w},\mathbf{z_1}) -  \mathbf{G}(\mathbf{w},\mathbf{z_2}) ||_1/||\mathbf{z_1}-\mathbf{z_2}||_1].
\end{equation}
$\mathcal{L}_{\texttt{div}}$ encourages the  $\mathbf{G}$ to diversify the synthesized feature space and enhance chances of generating features from minor modes.

\subsection{Unseen Synthesis and Detection}
Optimizing the loss defined in Eq.~\ref{eqn: full_gan_loss} results in conditional visual feature generator $\mathbf{G}$. We can synthesize an arbitrarily large number of features $\mathbf{\tilde{f}}_u=\mathbf{G}(\mathbf{z},\mathbf{w})$ for each unseen class by using its corresponding class semantics vector $\mathbf{w} \in \mathbf{W}_u$ and a random noise vector $\mathbf{z} \sim \mathcal{N}(\mathbf{0},\mathbf{1})$. Repeating the process for all unseen classes, we get synthesized features $\mathbf{\tilde{F}}_u$ and their corresponding class-labels $\mathbf{Y}_u$, which can then be used to update softmax classifier $\mathbf{\phi}_{\texttt{cls}}$ of $\phi_{\texttt{faster-rcnn}}$ for unseen classes. 
At inference, a simple forward pass through $\phi_{\texttt{faster-rcnn}}$ predicts both class-wise confidence scores and offsets for the bounding-box coordinates. We consider a fixed number of proposals from the RPN ($100$ in our case) and apply non-maximal suppression (NMS) with a threshold of $0.5$ to obtain final detections. The classification confidence for the proposals are directly given by $\mathbf{\phi}_{\texttt{cls}}$, whereas the bounding-box offset coordinates of an unseen class are estimated by the predictions for the seen class with maximum classification response. We observe that this is a reasonable assumption since visual features for the unseen class and its associated confusing seen class are similar. For the case of Generalized zero-shot-detection (GZSD), we simply consider all detections from seen and unseen objects together, whereas for ZSD, detections corresponding to seen objects are only considered.


\section{Results}

\noindent\textbf{Datasets:} We extensively evaluate our proposed ZSD method on three popular object detection datasets: MSCOCO 2014 \cite{MSCOCO_2014}, ILSVRC Detection 2017 \cite{ILSVRC_2015} and PASCAL VOC 2007/2012 \cite{VOC_IJCV_2010}. For MSCOCO, we use 65/15 seen/unseen split proposed in \cite{rahman2018polarity}. As argued in \cite{rahman2018polarity}, this split exhibits rarity and diverseness 
of the unseen classes in comparison to another 48/17 split proposed in \cite{bansal2018zero}. 
We use 62,300 images for training set and 10,098 images from the validation set for testing ZSD and GZSD. For ILSVRC Detection 2017, we follow the 177/23 seen/unseen split proposed in \cite{rahman2018zero} that provides 315,731 training images and 19,008 images for testing. For PASCAL VOC 2007/2012, we follow the 16/4 seen/unseen split proposed in \cite{demirel2018zero} that uses a total of 5,981 images from the train set of 2007 and 2012 and 1,402 images for testing from val+test set of PASCAL VOC 2007. To test the seen detection results, it uses 4,836 images from the test+val set of 2007. For all these datasets, the testing set for ZSD contains at least one unseen object per image.

\noindent\textbf{Implementation details:} We rescale each image to have the smaller side of {600, 800 and 600} pixels respectively for PASCAL VOC, MSCOCO and ILSVRC Detection datasets. For training our generative module, we consider different anchor bounding boxes with an IoU $\ge 0.7$ as foregrounds, whereas IoU $\le 0.3$ boxes are considered as background. We ignore other bounding-boxes with an IoU between 0.3 and 0.7, since a more accurate bounding box helps GAN in learning discriminative features. We first train our Faster-RCNN model on seen data for {12} epochs using standard procedure as in \cite{mmdetection}. Our category classifier $\phi_{\texttt{cls}}$, and bounding-box regressor $\phi_{\texttt{reg}}$ both have a single fully-connected layer. The trained model is then used to extract visual features corresponding to bounding-boxes of ground-truth seen objects. We then train our generative model to learn the underlying data distribution of the extracted seen visual features. 

The generator $\mathbf{G}$ and discriminator $\mathbf{D}$ of our GAN model are simple single-layered neural networks with {4096} hidden units. Through out our experiments, the loss re-weighting hyper-parameters in Eq.~\ref{eqn: full_gan_loss} are set as, $\alpha_1=1.0,  \alpha_2=0.1, \alpha_3=0.1, \alpha_4=1.0$, using a small held-out validation set. The noise vector $\mathbf{z}$ has the same dimensions as the class-semantics vector $\mathbf{w} \in \mathbb{R}^d$ and is drawn from a unit Gaussian distribution with zero mean. We use $\lambda=10$ as in \cite{gulrajani2017improved}. For training of our  cWGAN model, we use Adam optimizer with learning rate $10^{-4}$, $\beta_1=0.5, \beta_2=0.999$. The loss term $\mathcal{L}_{C_u}$ is included after first $5$ epochs, when the generator $\mathbf{G}$ has started to synthesize meaningful features. Once the generative module is trained, we synthesize 300 features for each unseen class, conditioned upon their class-semantics, and use them to train $\phi_{\texttt{cls}}$ for {30} epochs using {Adam} optimizer. To encode class-labels, unless mentioned otherwise, we use the FastText \cite{mikolov2018advances} embedding vectors learnt on large corpus of non-annotated text. 
The implementation of the proposed method in Pytorch is available at \url{https://github.com/nasir6/zero_shot_detection}

\noindent\textbf{Evaluation metrics:} Following previous works \cite{rahman2018polarity,bansal2018zero}, we report recall@100 (RE) and mean average precision (mAP) with IoU=0.5. We also report per-class average prevision (AP) to study category-wise performance. For GZSD, we report Harmonic Mean (HM) of performances for seen and unseen classes.


\subsection{Comparisons with the State-of-the-Art}

\subsubsection{Comparison methods:} We compare our method against a number of recently proposed state-of-the-art ZSD and GZSD methods. These include: \textbf{(a)} \textbf{SB}, \textbf{LAB} \cite{bansal2018zero}, which is a background-aware approach that considers external annotations from object instances belonging to neither seen or unseen. This extra information helps SB, LAB \cite{bansal2018zero} to address the confusion between unseen and background. \textbf{(b)} \textbf{DSES} \cite{bansal2018zero} is a version of above approach that does not use background-aware representations but employs external data sources for background. \textbf{(c)} \textbf{HRE} \cite{demirel2018zero}: A YOLO based end-to-end ZSD approach based on the convex combination of region embeddings. \textbf{(d)} \textbf{SAN} \cite{rahman2018zero}: A Faster-RCNN based ZSD approach that takes advantage of super-class information and a max-margin loss to understand unseen objects better. \textbf{(e)} \textbf{PL-48}, \textbf{PL-65} \cite{rahman2018polarity}: A RetinaNet based ZSD approach that uses polarity loss for better alignment of visual features and semantics. 
\textbf{(f)} \textbf{ZSDTD} \cite{li2019zero}: This approach uses textual description instead of a single-word class-label to define semantic representation. The additional textual description enriches the semantic space and helps to better relate semantics with the visual features. \textbf{(g)} \textbf{GTNet} \cite{zhao2020gtnet}: uses multiple GAN models alongwith textual descriptions similar to \cite{li2019zero}, to generate unseen features to train a Faster-RCNN based ZSD model in a supervised manner. \textbf{(h)} \textbf{Baseline}: The baseline method trains a standard Faster-RCNN model for seen data $\mathcal{X}^s$. To extend it to unseen classes for ZSD, it first gets seen predictions $\mathbf{p}_s$, and then project them onto class semantics to get unseen predictions $\mathbf{p}_u = \mathbf{W}_u\mathbf{W}_s^{T}\mathbf{p}_s$ as in \cite{rahman2018polarity}. (i) \textbf{Ours}: This is our proposed ZSD approach.

\begin{table}[!t]
\caption{\small ZSD and GZSD performance of different methods on MSCOCO in terms of mAP and recall (RE). Note that our proposed feature synthesis based approach achieves a significant gain over the existing state-of-the-art. For the mAP metric, compared with the second best method PL-65 \cite{rahman2018polarity}, our method shows a relative gain of $53\%$ on ZSD and $38\%$ on harmonic mean of seen and unseen for GZSD.
}
\centering\setlength{\tabcolsep}{4pt}
\scalebox{0.9}{
\begin{tabular}{c|c|c|c|c|c|c}
\toprule[0.1em]
\rowcolor{mygray}
&  & Seen/Unseen &  & \multicolumn{3}{|c}{GZSD} \\ \cline{5-7}
\rowcolor{mygray} \multirow{-2}{*}{Metric} & \multirow{-2}{*}{Method}  & split 	& \multirow{-2}{*}{ZSD} & seen	& unseen & HM   \\ \toprule[0.1em]
\multirow{7}{*}{mAP}&SB~\cite{bansal2018zero} &48/17&0.70 &-&-&- \\
&DSES~\cite{bansal2018zero} &48/17&0.54&-&-&- \\
&PL-48 \cite{rahman2018polarity} &48/17&10.01 &35.92 &4.12 &7.39 \\
&PL-65 \cite{rahman2018polarity}  & 65/15  & 12.40 & 34.07 & 12.40 & 18.18 \\
\cline{2-7} 
&Baseline & 65/15 &8.80 & 36.60 & 8.80 & 14.19 \\
&Ours & 65/15 & \textbf{19.0} & \textbf{36.90}  &\textbf{19.0} & \textbf{25.08} \\
\hline\hline

\multirow{7}{*}{RE}&SB~\cite{bansal2018zero} &48/17&24.39&-&-&- \\
&DSES~\cite{bansal2018zero} &48/17&27.19&15.02&15.32&15.17 \\
&PL-48 \cite{rahman2018polarity} &48/17&43.56&38.24&26.32&31.18 \\
&PL-65 \cite{rahman2018polarity}  & 65/15  & 37.72 & 36.38 & 37.16 & 36.76 \\
\cline{2-7}  
& Baseline & 65/15 & 44.40 & 56.40& 44.40 & 49.69 \\
& Ours & 65/15 & \textbf{54.0} & \textbf{57.70} & \textbf{53.90} & \textbf{55.74} \\ 
\bottomrule[0.1em]
\end{tabular}}
\vspace{0.5em}  

\label{tab:mscoco_map}
\end{table}

\begin{table}[!t]
  \caption{\small Class-wise AP comparison of different methods on unseen classes of MSCOCO for ZSD. The proposed method shows significant gains for a number of individual classes. Compared with the second best method PL \cite{rahman2018polarity}, our method shows an absolute mAP gain of $6.6\%$.} 
  \centering\setlength\tabcolsep{2.6pt}
  \scalebox{0.82}{
    \begin{tabular}{c|c|c|c|c|c|c|c|c|c|c|c|c|c|c|c|c}
   \toprule[0.1em]
\rowcolor{mygray}
{\textbf{{Method}}}&\rotatebox{90}{\textbf{Overall}}&\rotatebox{90}{{aeroplane}}&\rotatebox{90}{{train}}&\rotatebox{90}{{parking}}\rotatebox{90}{{ meter}}&\rotatebox{90}{{cat}}&\rotatebox{90}{{bear}}&\rotatebox{90}{{suitcase}}&\rotatebox{90}{{frisbee}}&\rotatebox{90}{{snow- }}\rotatebox{90}{{board}}&\rotatebox{90}{{fork}}&\rotatebox{90}{{sand-}}\rotatebox{90}{{wich}}&
\rotatebox{90}{{hot dog}}&\rotatebox{90}{{toilet}}&\rotatebox{90}{{mouse}}&\rotatebox{90}{{toaster}}&\rotatebox{90}{{hair }} \rotatebox{90}{{drier}} \\
\toprule[0.1em]
PL-Base \cite{rahman2018polarity}  & 8.48 &4.0 &28.7& .29 &18.0& 0.0 &13.1 &11.3& 24.3 &13.8& 9.6& 2.0& 1.1 &.24 &.73& 0.0 \\
 PL \cite{rahman2018polarity} & 12.40  &20.0 & 48.2 & .63 & 28.3  &13.8  &12.4 & 21.8 & 15.1 & 8.9 & 8.5  &.87 & 5.7 & .04  &1.7 & .03\\
Ours-Baseline & 8.80 & 1.9&31.8&0.0&59.3&3.8&0.6&0.1&19.6&10.7&2.8&0.0&0.8&0.0&0.0&0.0 \\

\hline
Ours & \textbf{19.0} &10.1&\textbf{48.7}&\textbf{1.2}&\textbf{64.0}&\textbf{64.1}&12.2&0.7&\textbf{28.0}&\textbf{16.4}&\textbf{19.4}&0.1&\textbf{18.7}&\textbf{1.2}&0.5&\textbf{0.2}\\ \bottomrule[0.1em]
    \end{tabular}}\vspace{0.5em}

  \label{tab:coco_class_wise}
\end{table}


\subsubsection{MSCOCO results:} Our results and comparisons with different state-of-the-art methods for ZSD and GZSD on MSCOCO dataset are presented in Table \ref{tab:mscoco_map}. \\
\emph{(a) ZSD results:} The results demonstrate that our proposed method achieves a significant gain on both metrics (mAP and RE) over the existing methods on ZSD setting. The gain is specifically  pronounced for the mAP metric, which is more challenging and meaningful to evaluate object detection algorithms. This is because mAP penalizes false positives while the RE measure does not impose any penalty on such errors. Despite the challenging nature of mAP metric, our method achieves a relative mAP gain of $53\%$ over the second-best method (PL~\cite{rahman2018polarity}). We attribute such remarkable improvement to the fact that our approach addresses the zero shot learning problem by augmenting the visual features. In contrast, previous approaches such as SB \cite{bansal2018zero}, DSES \cite{bansal2018zero}, PL \cite{rahman2018polarity} map visual features to the semantic space that limits their flexibility to learn strong representations mainly due to the noise in semantic domain. In comparison, our approach helps in reducing the biases towards the seen classes during training, avoids unseen-background confusion, and minimizes the hubness problem. 

\begin{figure}[tp]
    \centering
    \includegraphics[trim=0cm 0.3cm 0cm 0cm, clip=true,width=0.7 \textwidth]{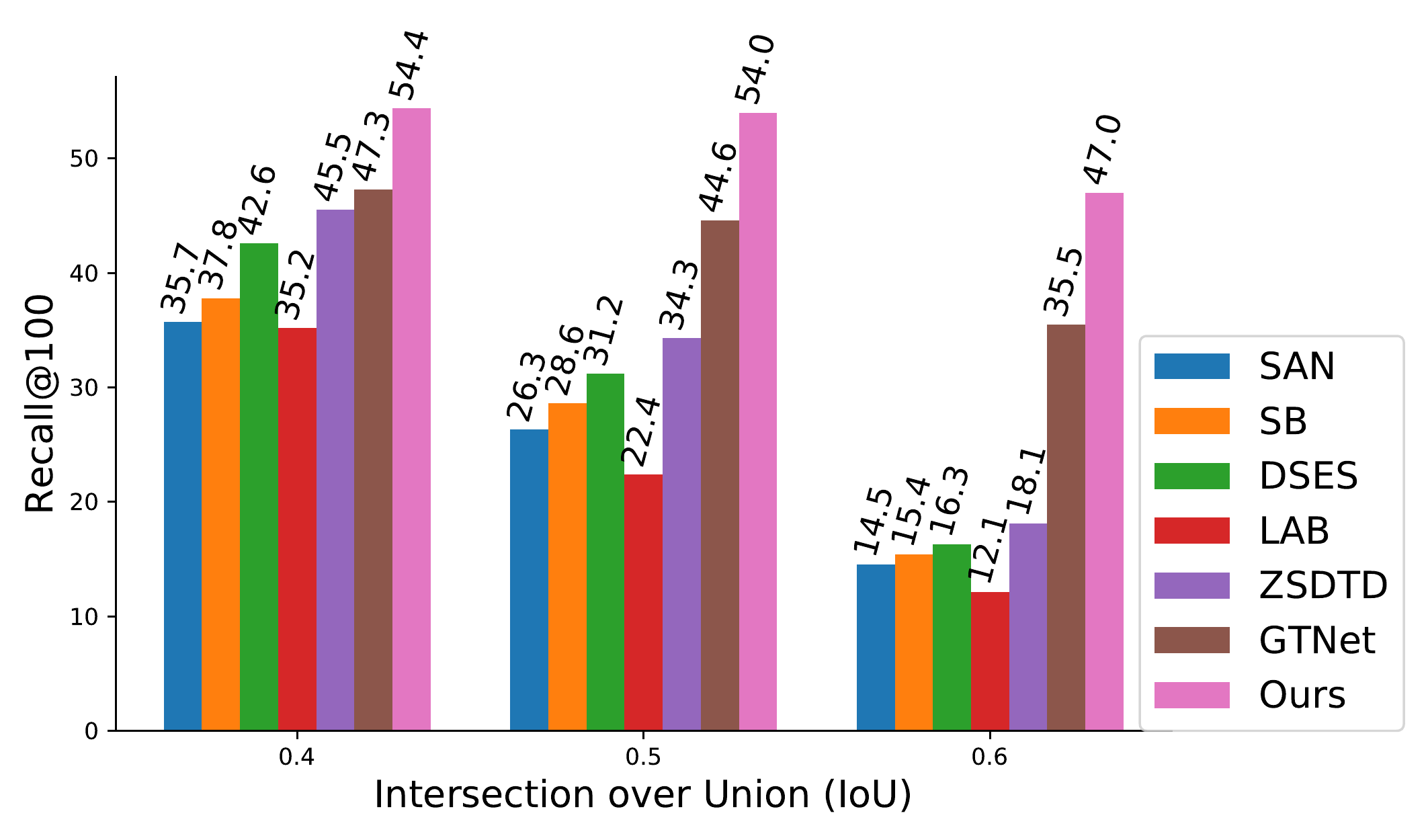}
    \caption{Comparison of SAN \cite{rahman2018zero}, SB/DES/LAB \cite{bansal2018zero}, ZSDTD \cite{li2019zero}, GTNet \cite{zhao2020gtnet} in terms of Recall@100 rates for different IoU settings on MSCOCO dataset. The proposed method consistently shows improved performance over existing state-of-the-art methods.}
    \label{fig:iou_coco}
\end{figure}

In Fig.~\ref{fig:iou_coco}, we further show comparisons for ZSD recall@100 rates by varying the IoU. Note that the compared methods in Fig.~\ref{fig:iou_coco} use additional information in the form of textual description of concepts instead of a single-word class name. Even though, our proposed method uses much simpler semantic information (only semantic vectors for class labels), the results in Fig.~\ref{fig:iou_coco} indicate that our method consistently outperforms several established methods by a large margin for a variety of IoU settings. This comparison includes a recent generative ZSD approach, GTNet~\cite{zhao2020gtnet}, that employs an ensemble of GANs to synthesize features. \\

\noindent\emph{(b) GZSD results:} Our GZSD results in Table~\ref{tab:mscoco_map} also achieve a significant boost in performance. The generated synthesized features allow training of the detection model in a supervised manner. In this way, unseen instances get equal emphases as seen class objects during training. We note that the GZSD setting is more challenging and realistic since both seen and unseen classes are present at inference. An absolute HM mAP gain of $6.9\%$ for GZSD is therefore quite significant for our proposed method.

Compared with the baseline, which projects visual features to semantic space, our results demonstrate the effectiveness of augmenting the visual space, and learning a discriminative classifier for more accurate classification. These baseline results further indicate the limitations of mapping multiple visual features to a single class-semantic vector.  One interesting trend is that the baseline still performs reasonably well according to the RE measure (in some cases even above the previous best methods), however the considerably low mAP scores tell us that the inflated performance from the baseline is prone to many false positives, that are not counted in the RE measure. For this reason, we believe the mAP scores are a more faithful depiction of ZSD methods.\\

\noindent\emph{(c) Class-wise performances:} Our class-wise AP results on MSCOCO in Table~\ref{tab:coco_class_wise} show that the performance gain for the proposed method is more pronounced for `\emph{train}', `\emph{bear}' and `\emph{toilet}' classes. Since our feature generation is conditioned upon class-semantics, we observe that the feature synthesis module generates more meaningful features for unseen classes which have similar semantics in the seen data. The method shows worst performance for classes `\emph{parking-meter}', `\emph{frisbee}', `\emph{hot dog}' and `\emph{toaster}'. These classes do not have close counterparts among the seen classes, which makes their detection harder. \\ 

\begin{figure}[tp]
    \centering
    \includegraphics[width=.192\textwidth,height=.15\textwidth]{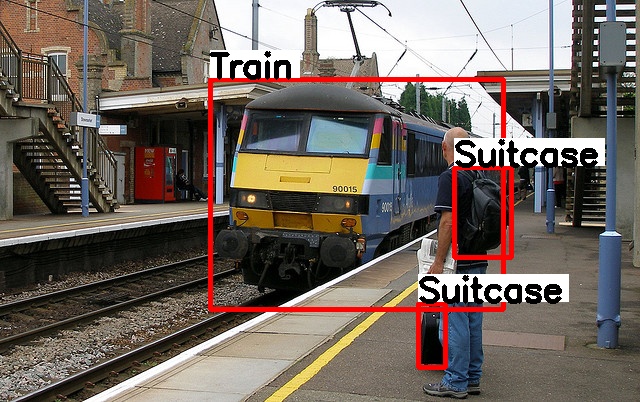}
    \includegraphics[width=.192\textwidth,height=.15\textwidth]{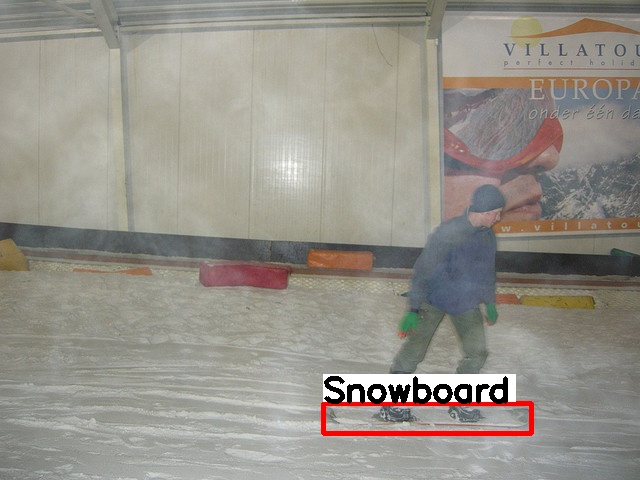}
    \includegraphics[width=.192\textwidth,height=.15\textwidth]{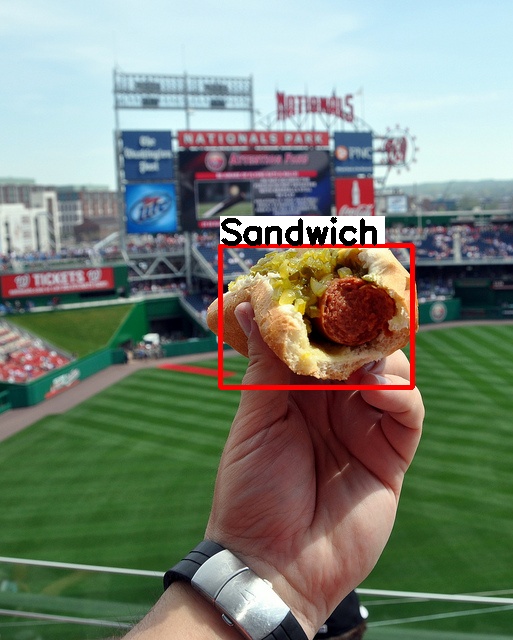}
    \includegraphics[width=.192\textwidth,height=.15\textwidth]{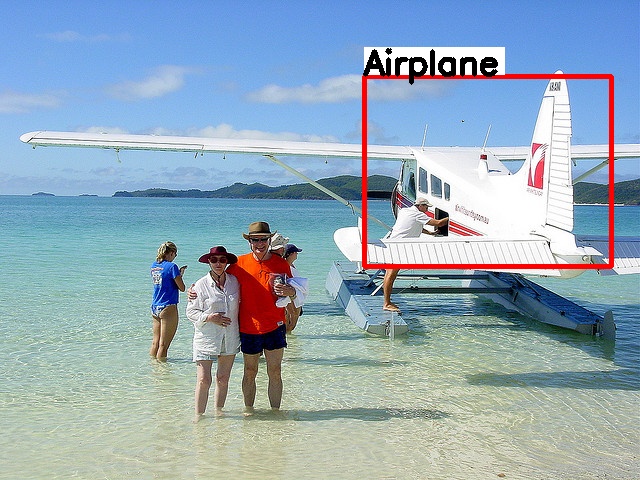}
    \includegraphics[width=.192\textwidth,height=.15\textwidth]{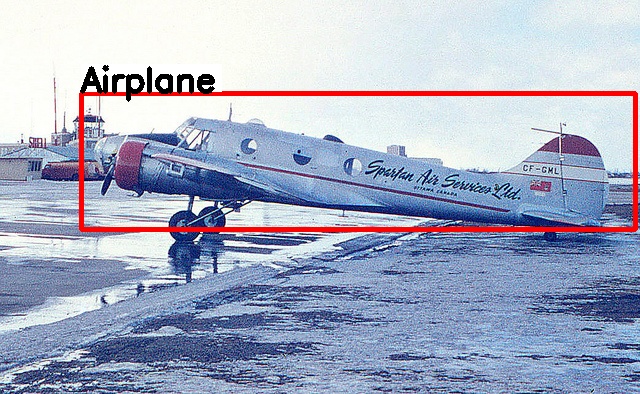} \\
    
    \includegraphics[width=.192\textwidth,height=.15\textwidth]{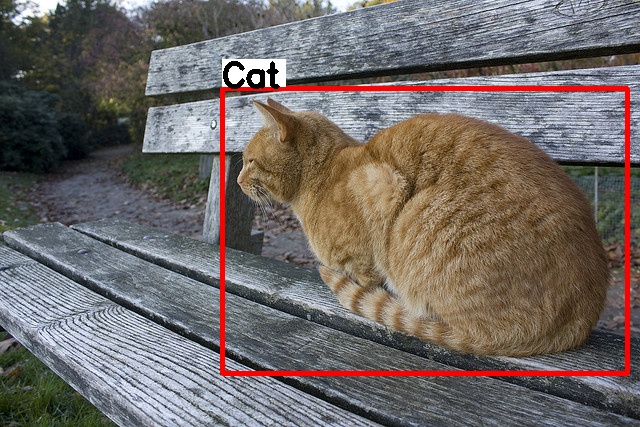}
    \includegraphics[width=.192\textwidth,height=.15\textwidth]{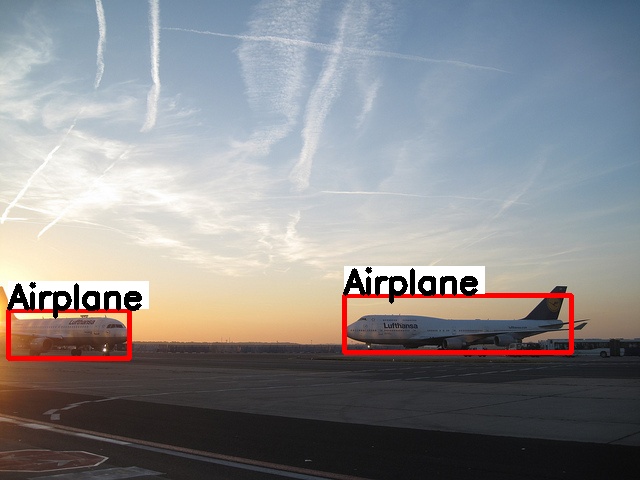}
    \includegraphics[width=.192\textwidth,height=.15\textwidth]{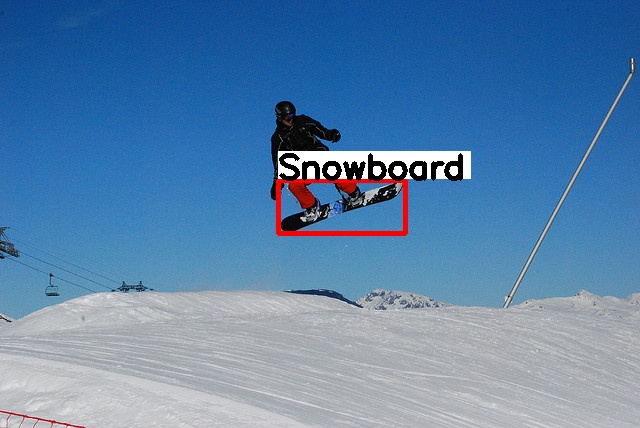}
    \includegraphics[width=.192\textwidth,height=.15\textwidth]{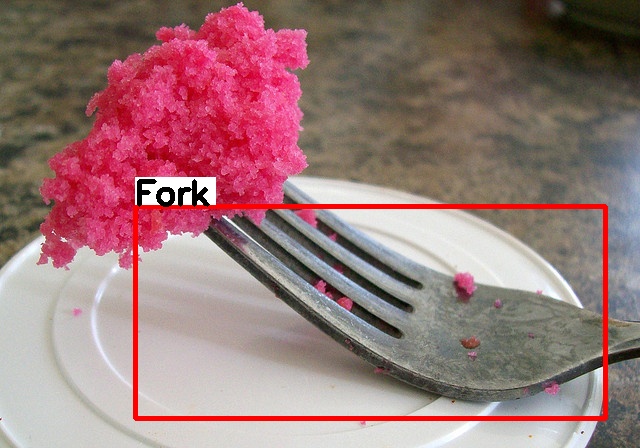}
    \includegraphics[width=.192\textwidth,height=.15\textwidth]{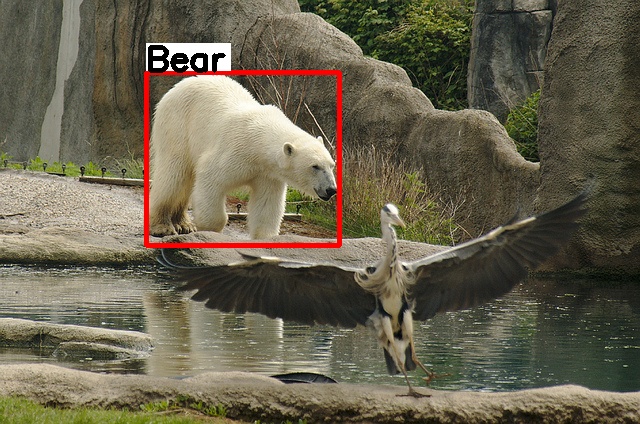}   \\
    
    Zero Shot Detection (ZSD) \\  \vspace{.5em}
    
    \includegraphics[width=.192\textwidth,height=.15\textwidth]{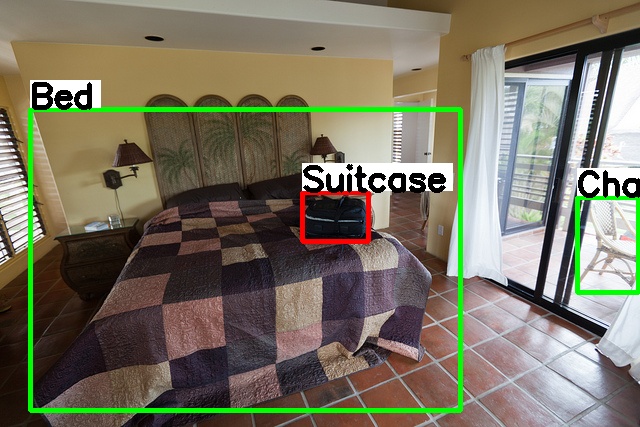}
    \includegraphics[width=.192\textwidth,height=.15\textwidth]{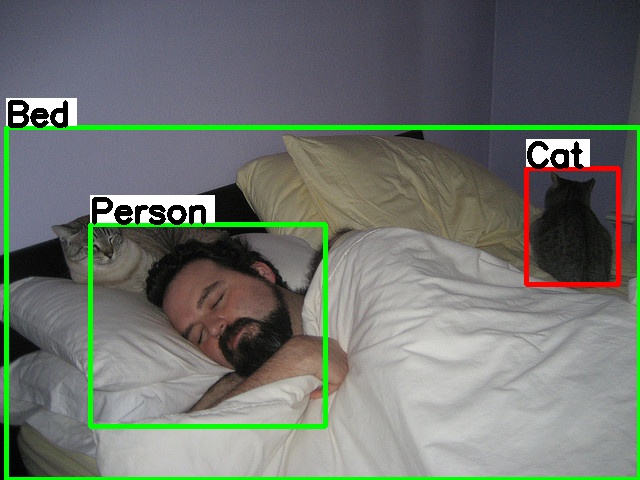}
    \includegraphics[width=.192\textwidth,height=.15\textwidth]{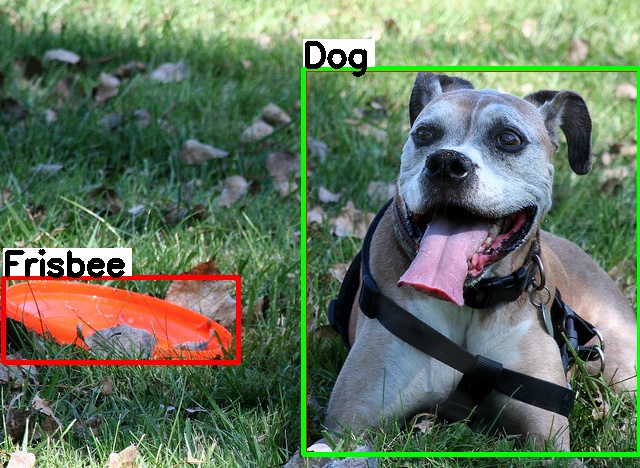}
    \includegraphics[width=.192\textwidth,height=.15\textwidth]{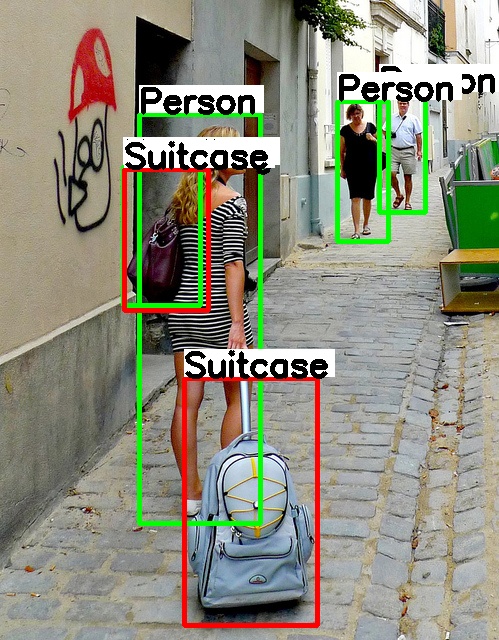}
    \includegraphics[width=.192\textwidth,height=.15\textwidth]{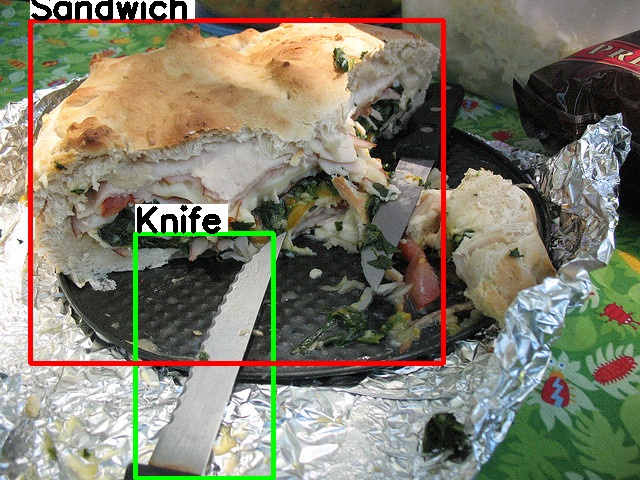} \\
    
    \includegraphics[width=.192\textwidth,height=.15\textwidth]{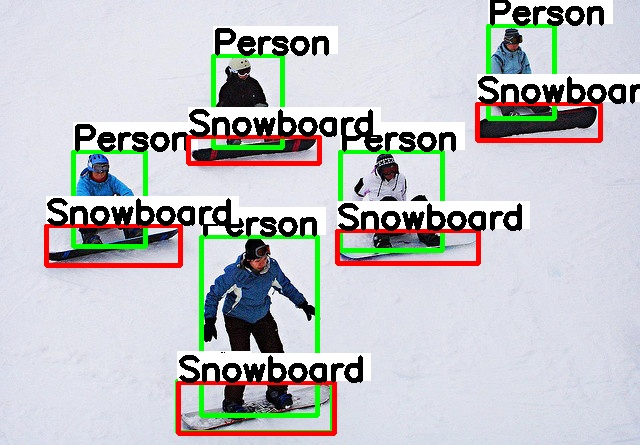}
    \includegraphics[width=.192\textwidth,height=.15\textwidth]{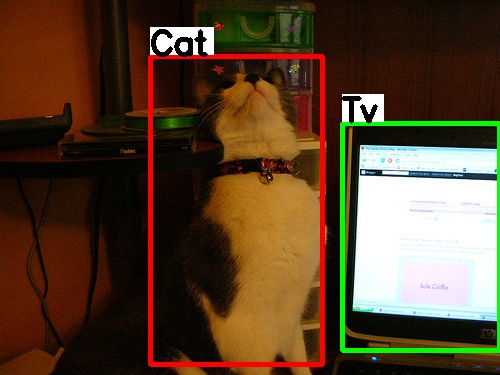}
    \includegraphics[width=.192\textwidth,height=.15\textwidth]{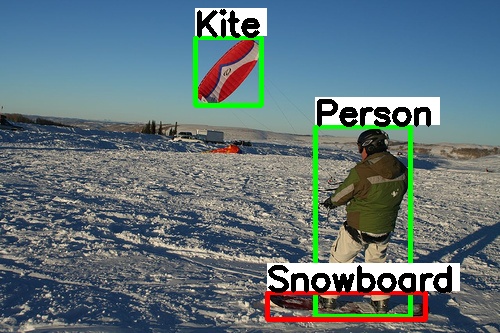}
    \includegraphics[width=.192\textwidth,height=.15\textwidth]{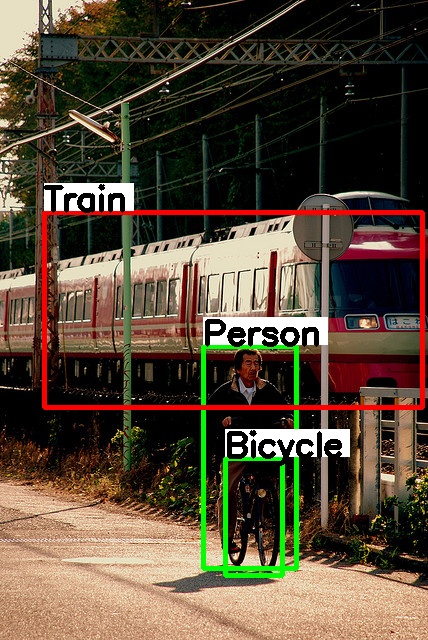}
    \includegraphics[width=.192\textwidth,height=.15\textwidth]{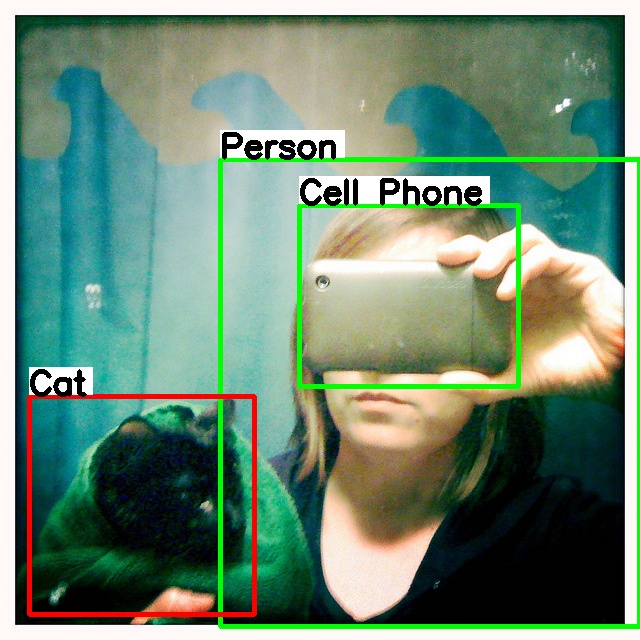}   
    
    Generalized Zero Shot Detection (GZSD)\\  
    
    \caption{Qualitative results on MSCOCO for ZSD (top 2 rows) and GZSD (bottom 2 rows). Seen classes are shown with green and unseen with red. \emph{(best seen when zoomed)}}
    \label{fig: qual}
\end{figure}

\noindent\emph{(d) Qualitative results:} Fig.~\ref{fig: qual} shows some examples of detections from our method both for ZSD (top 2 rows) and GZSD (bottom 2 rows) settings. The visual results demonstrate the effectiveness of the proposed method in localizing unseen objects, and its capability to detect multiple seen+unseen objects with challenging occlusions and background clutter in real-life images.

\subsubsection{PASCAL VOC results:} 
\begin{figure}[tp]
    \centering
    \fbox{
    \includegraphics[width=.312\textwidth, height=.25\textwidth]{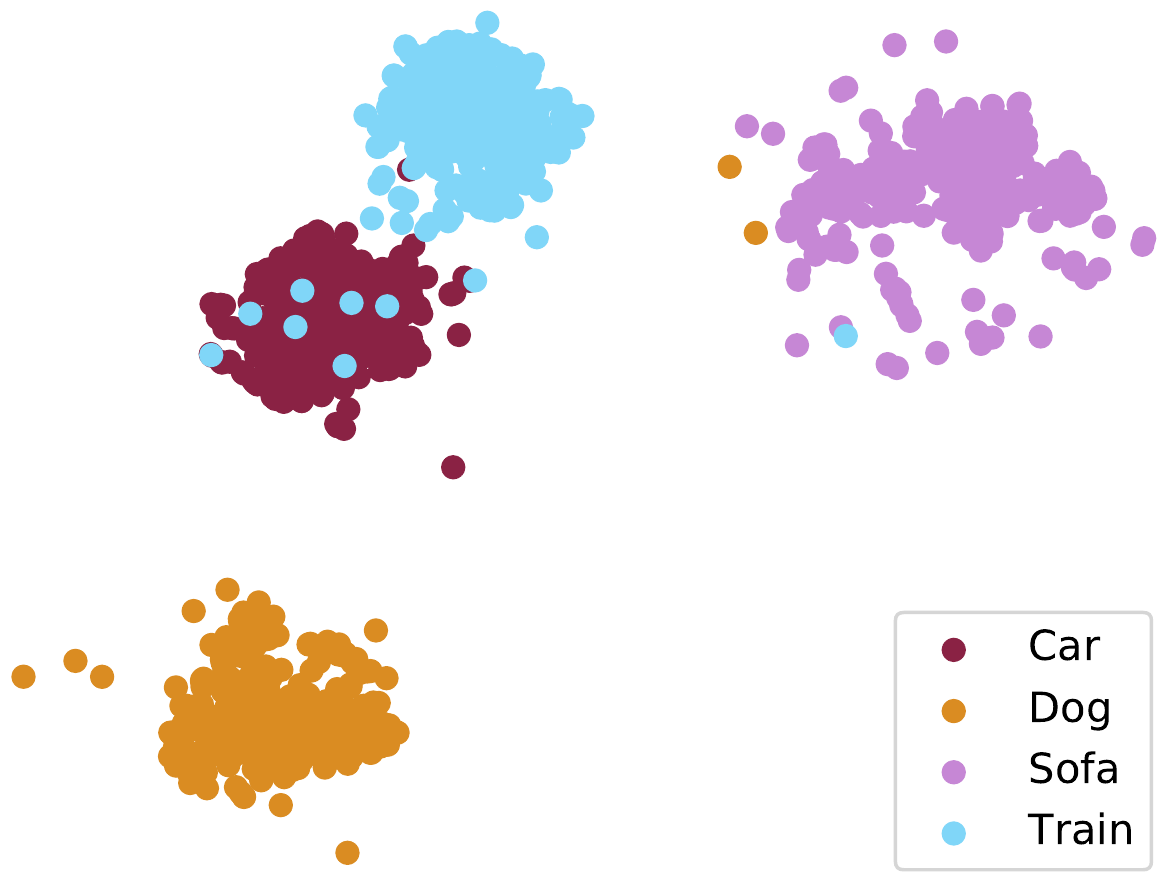}} \hspace{2em}
    \fbox{
    \includegraphics[width=.45\textwidth, height=.25\textwidth]{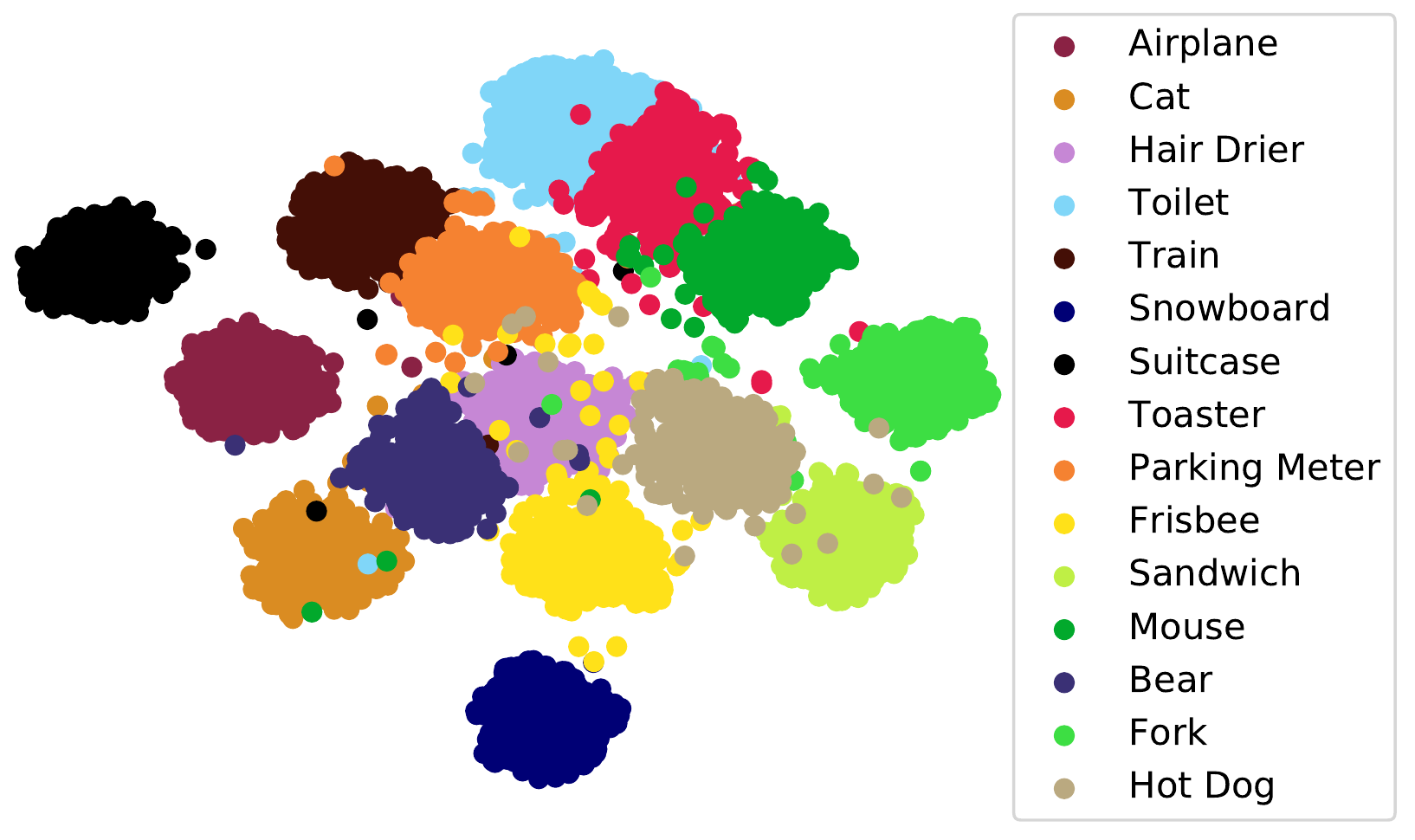}
    }
    
    \caption{A t-SNE visualization of synthesized features by our approach for unseen classes on PASCAL VOC dataset (left) and MSCOCO dataset (right). The generated features form well-separated and distinctive clusters for different classes.}
    \label{fig: tsne}
\end{figure}

\begin{figure}
    \centering
    \includegraphics[width=.192\textwidth,height=.14\textwidth]{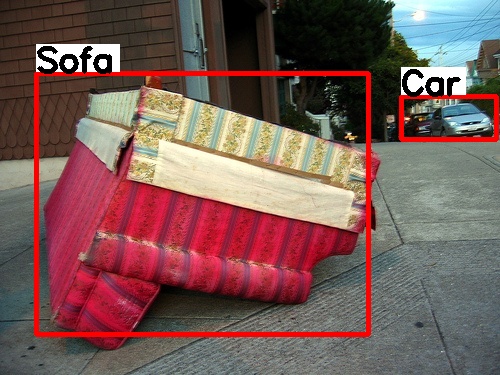}
    \includegraphics[width=.192\textwidth,height=.14\textwidth]{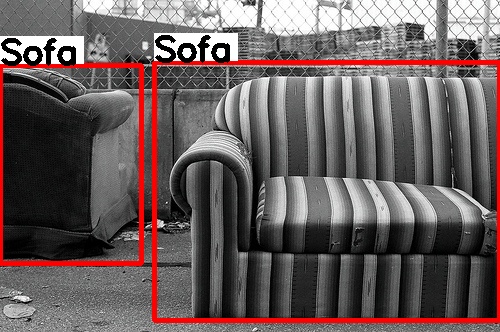}
    \includegraphics[width=.192\textwidth,height=.14\textwidth]{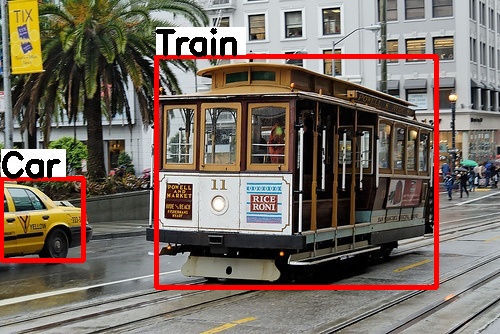}
    \includegraphics[width=.192\textwidth,height=.14\textwidth]{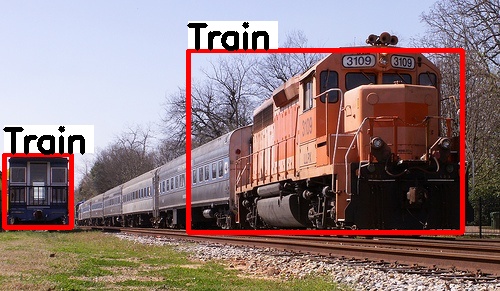}
    \includegraphics[width=.192\textwidth,height=.14\textwidth]{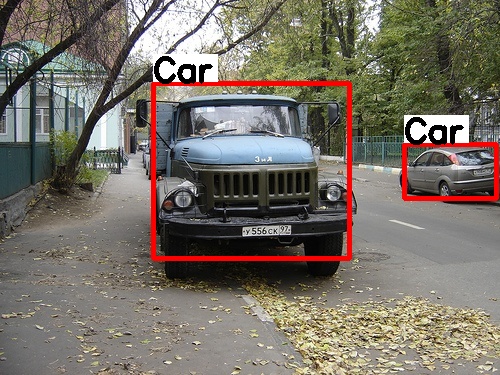} \\
    
    \includegraphics[width=.192\textwidth,height=.14\textwidth]{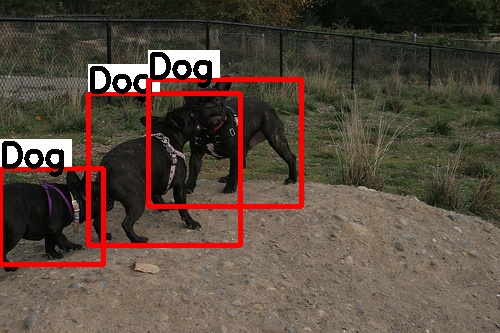}
    \includegraphics[width=.192\textwidth,height=.14\textwidth]{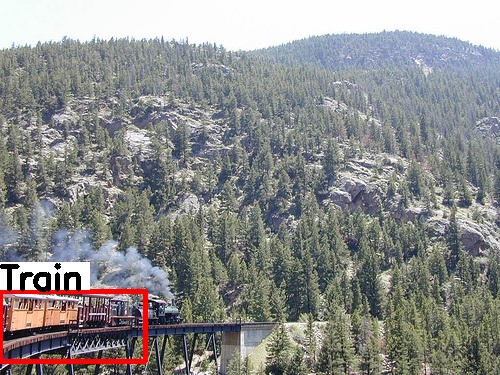}
    \includegraphics[width=.192\textwidth,height=.14\textwidth]{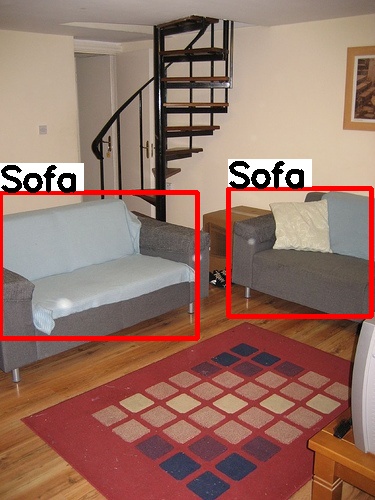}
    \includegraphics[width=.192\textwidth,height=.14\textwidth]{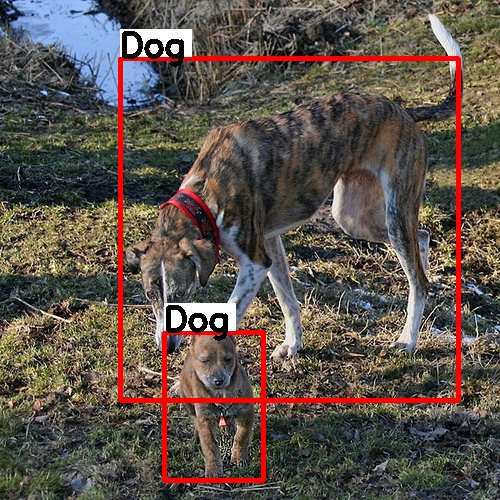}
    \includegraphics[width=.192\textwidth,height=.14\textwidth]{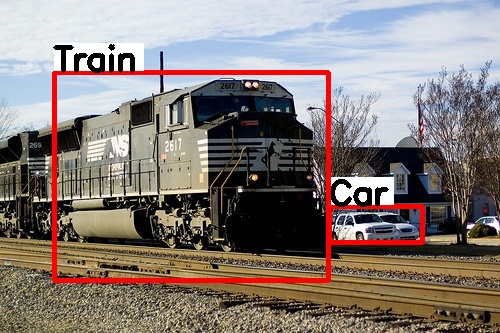}   \\

    \caption{Example unseen detections on PASCAL VOC. \emph{(best seen when zoomed)}.}
    \label{fig: qual_voc}
\end{figure}

In Table \ref{tab:PASCAL VOC}, we compare different methods on PASCAL VOC dataset based on the setting mentioned in \cite{demirel2018zero}. The results suggest that the proposed method achieves state-of-the-art ZSD performance. A few examples images on PASCAL VOC shown in Fig.~\ref{fig: qual_voc} demonstrate the capability of our method to detect multiple unseen objects in real-life scenarios. The results in Table \ref{tab:PASCAL VOC} further indicate that in addition to the unseen detection case, our method performs very well in the traditional seen detection task. We outperform the current best model PL \cite{rahman2018polarity} by a significant margin, i.e., $73.6\%$ vs. $63.5\%$ for seen detection and $64.9\%$ vs. $62.1\%$ for unseen detection. A t-SNE visualization of our synthesized features for unseen classes is shown in Fig.~\ref{fig: tsne}. We observe that our generator can effectively capture the underlying data distribution of visual features. The similar classes occur in close proximity of each other. We further observe that the synthesized features form class-wise clusters that are distinctive, thus aiding in learning a discriminative classifier on unseen classes. Synthesized features for similar classes (\emph{bus} and \emph{train}) are however sometimes confused with each other due to high similarity in their semantic space representation.

\begin{table}[!t]
  \caption{  mAP scores on PASCAL VOC'07. \underline{\textit{Italic}} classes are unseen.} 

  \centering\setlength\tabcolsep{0.7pt}
  \scalebox{.7}{
    \begin{tabular}{c|c|c||c|c|c|c|c|c|c|c|c|c|c|c|c|c|c|c||c|c|c|c}
    
\toprule[0.1em]
\rowcolor{mygray}
  
\rotatebox{90}{\textbf{Method}}&\rotatebox{90}{{ \textbf{Seen}}}&\rotatebox{90}{{ \textbf{Unseen}}}&\rotatebox{90}{aeroplane}&\rotatebox{90}{bicycle}&\rotatebox{90}{bird}&\rotatebox{90}{boat}&\rotatebox{90}{bottle}&\rotatebox{90}{bus}&\rotatebox{90}{cat}&\rotatebox{90}{chair}&\rotatebox{90}{cow}&\rotatebox{90}{d.table}&\rotatebox{90}{horse}&\rotatebox{90}{motrobike}&\rotatebox{90}{person}&\rotatebox{90}{p.plant}&\rotatebox{90}{sheep}&\rotatebox{90}{tvmonitor}&\rotatebox{90}{{\underline{\textit{car}}}}&\rotatebox{90}{\underline{\textit{dog}}}&\rotatebox{90}{\underline{\textit{sofa}}}&\rotatebox{90}{\underline{\textit{train}}} \\  
\toprule[0.1em]
HRE \cite{demirel2018zero}&57.9&54.5&68.0&72.0&74.0&48.0&41.0&61.0&48.0&25.0&48.0&73.0&75.0&71.0&73.0&33.0&59.0&57.0&55.0&82.0&55.0&26.0 \\ 
PL \cite{rahman2018polarity} &63.5&62.1&74.4&71.2&67.0&50.1&50.8&67.6&84.7&44.8&68.6&39.6&74.9&76.0&79.5&39.6&61.6&66.1&63.7&87.2&53.2&44.1 \\ \hline

Ours &\textbf{73.6}&\textbf{64.9}&\textbf{83.0}&\textbf{82.8}&\textbf{75.1}&\textbf{68.9}&\textbf{63.8}&\textbf{69.5}&
\textbf{88.7}&\textbf{65.1}&\textbf{71.9}&{56.0}&\textbf{82.6}&\textbf{84.5}&\textbf{82.9}&\textbf{53.3}&\textbf{74.2}&\textbf{75.1}&59.6&\textbf{92.7}&\textbf{62.3}&\textbf{45.2} \\ 
\bottomrule[0.1em]
    \end{tabular}}\vspace{0.5em}
  \label{tab:PASCAL VOC}
\end{table}

\begin{table}[!t]
  \caption{ZSD class-wise AP for unseen classes of ILSVRC DET 2017 dataset.} 

  \begin{center}\setlength\tabcolsep{2pt}
  \scalebox{0.6}{
    \begin{tabular}{c|c||c|c|c|c|c|c|c|c|c|c|c|c|c|c|c|c|c|c|c|c|c|c|c}
    \toprule[0.1em]
\rowcolor{mygray}
    
\rotatebox{90}{ }&\rotatebox{90}{ \textbf{mean}}&\rotatebox{90}{p.box}&\rotatebox{90}{syringe}&\rotatebox{90}{harmonica}&\rotatebox{90}{maraca}&\rotatebox{90}{burrito}&\rotatebox{90}{pineapple}&\rotatebox{90}{electric-fan}&\rotatebox{90}{iPod}&\rotatebox{90}{dishwasher}&\rotatebox{90}{canopener}&\rotatebox{90}{plate-rack}&\rotatebox{90}{bench}&\rotatebox{90}{bowtie}&\rotatebox{90}{s.trunk}&\rotatebox{90}{scorpion}&\rotatebox{90}{snail}&\rotatebox{90}{hamster}&\rotatebox{90}{tiger}&\rotatebox{90}{ray}&\rotatebox{90}{train}&\rotatebox{90}{unicycle}&\rotatebox{90}{golfball}&\rotatebox{90}{h.bar}\\  \toprule[0.1em]
Baseline &12.7 &0.0 &3.9 &0.5 &0.0& 36.3& 2.7& 1.8 &1.7& 12.2& 2.7& 7.0 &1.0 &0.6 &22.0 &19.0& 1.9 &40.9 &75.3& 0.3 &28.4& 17.9& 12.0 & 4.0 \\

SAN ($L_{mm}$)& 15.0& 0.0& 8.0 &0.2& 0.2 &39.2& 2.3 &1.9& 3.2& 11.7 &4.8& 0.0 &0.0& 7.1& 23.3 &25.7 &5.0 &50.5& 75.3& 0.0 &44.8& 7.8& 28.9& 4.5 \\
SAN\cite{rahman2018zero} &16.4 &5.6 &1.0 &0.1 &0.0 &27.8 &1.7& 1.5 &1.6 &7.2& 2.2& 0.0& 4.1& {5.3}& {26.7} &{65.6}& 4.0 &47.3& {71.5}& {21.5} &51.1& 3.7 &26.2& {1.2} \\
\hline
Ours & \textbf{24.3}&\textbf{6.2} &\textbf{18.6} & \textbf{0.7}&\textbf{5.9} &\textbf{50.9} &\textbf{8.2} &\textbf{2.1} &\textbf{55.3} &{11.5} &\textbf{14.3} &{3.0} &\textbf{15.4} &2.7 &11.4 &41.9&\textbf{16.4} & \textbf{79.6}&67.6 &14.5 &\textbf{69.5} &\textbf{31.8} &\textbf{30.7} &0.1  \\

\bottomrule[0.1em]

    \end{tabular}}
  \end{center}
  \label{tab:ilsvrc}
\end{table}

\subsubsection{ILSVRC DET 2017 results:} In Table \ref{tab:ilsvrc}, we report ZSD results on ILSVRC Detection dataset based on the settings mentioned in \cite{rahman2018zero}. We can notice from the results that, in most of the object categories, we outperform our closed competitor SAN~\cite{rahman2018zero} by a large margin. Note that for a fair comparison, we do not compare our method with reported results in \cite{li2019zero,zhao2020gtnet}, since both these methods use additional information in the form of textual description of class-labels. It has been previously shown in \cite{li2019zero} that the additional textual description information boosts performance across the board. For example, in their paper, SAN \cite{rahman2018zero} reports an mAP of $16.4$ using single-word description for class-labels, whereas, \cite{li2019zero} reports an mAP of $20.3$ for SAN using multi-word textual description of class-labels. Our improvements over SAN again demonstrates the significance of the proposed generative approach for synthesizing unseen features.



\subsubsection{Distinct Foreground Bounding Boxes:} The seen visual features are extracted based upon the anchor bounding boxes generated by using the ground-truth bounding boxes for seen classes in $\mathcal{X}^s$. We perform experiments by changing the definition of background and foreground bounding-boxes. Specifically, we consider two settings: \textbf{(a) }Distinct bounding-boxes: foreground object has a high overlap (IoU $\ge$ 0.7), and the background has minimal overlap with the object (IoU $\le$ 0.3), and  \textbf{(b)} Overlapping bounding-boxes: foreground has a medium overlap with the object of interest (IoU $>$ 0.5), and some background boxes have medium overlap with the object (IoU $<$ 0.5). We achieve an mAP of 19.0 vs 11.7 for distinct and overlapping boundig boxes respectively  on MSCOCO 65/15 split. This suggests that the generative module synthesizes the most discriminant features when the bounding-boxes corresponding to the real visual features have a high overlap with the respective object and minimal background.

\section{Conclusion}
The paper proposed a feature synthesis approach for simultaneous localization and categorization of objects in the framework of ZSD and GZSD. The proposed method can effectively learn the underlying visual-feature data distribution, by training a generative adversarial network model conditioned upon class-semantics. The GAN training is driven by a semantic-space unseen classifier, a seen classifier and a diversity enhancing regularizer. The method can therefore synthesize high quality unseen features which are distinct and discriminant for the subsequent classification stage. The proposed framework generalizes well to both seen and unseen objects and achieves impressive performance gains on a number of evaluated benchmarks including MSCOCO, PASCAL VOC and ILSVRC detection datasets. 

\bibliographystyle{splncs}
\bibliography{egbib}
\end{document}